\documentclass[conference, letterpaper]{IEEEtran}
\IEEEoverridecommandlockouts
\usepackage{cite}
\usepackage{amsmath,amssymb,amsfonts}
\usepackage{algorithmic}
\usepackage{graphicx}
\usepackage{textcomp}
\usepackage{xcolor}
\usepackage{hyperref}
\usepackage{makecell}
\usepackage{lipsum}
\usepackage{fancyhdr}
\usepackage{orcidlink}
\usepackage{mwe}
\hypersetup{
    colorlinks=true,
    linkcolor=blue,
    filecolor=magenta,      
    urlcolor=blue,
    pdftitle={Overleaf Example},
    pdfpagemode=FullScreen,
    }

\def\BibTeX{{\rm B\kern-.05em{\sc i\kern-.025em b}\kern-.08em
    T\kern-.1667em\lower.7ex\hbox{E}\kern-.125emX}}
\begin{document}

\fancyhf{}
\renewcommand{\headrulewidth}{0pt}
\fancyfoot[c]{}
\fancypagestyle{FirstPage}{
\lfoot{979-8-3503-7457-5/24/\$31.00 \copyright2024 IEEE} 
}

\title{Graph Learning-based Regional Heavy Rainfall Prediction Using Low-Cost Rain Gauges}

\author{\IEEEauthorblockN{Edwin Salcedo \orcidlink{0000-0001-8970-8838}} 
\IEEEauthorblockA{\textit{Department of Mechatronics Engineering} \\
\textit{Universidad Católica Boliviana ``San Pablo''}\\
    La Paz, Bolivia \\
    esalcedo@ucb.edu.bo 
    }
}

\maketitle

\begin{abstract}
Accurate and timely prediction of heavy rainfall events is crucial for effective flood risk management and disaster preparedness. By monitoring, analysing, and evaluating rainfall data at a local level, it is not only possible to take effective actions to prevent any severe climate variation but also to improve the planning of surface and underground hydrological resources. However, developing countries often lack the weather stations to collect data continuously due to the high cost of installation and maintenance. In light of this, the contribution of the present paper is twofold: first, we propose a low-cost IoT system for automatic recording, monitoring, and prediction of rainfall in rural regions. Second, we propose a novel approach to regional heavy rainfall prediction by implementing graph neural networks (GNNs), which are particularly well-suited for capturing the complex spatial dependencies inherent in rainfall patterns. The proposed approach was tested using a historical dataset spanning 72 months, with daily measurements, and experimental results demonstrated the effectiveness of the proposed method in predicting heavy rainfall events, making this approach particularly attractive for regions with limited resources or where traditional weather radar or station coverage is sparse.
\end{abstract}


\section{Introduction}
\thispagestyle{FirstPage}
\label{sec:introduction}

In past decades, the agricultural sector in Bolivia has experienced many ups and downs due to natural disasters, including floods, droughts, hail, and frost. Among these weather adversities, flooding causes considerable agricultural losses in Latin America and has been a frequent issue in rural areas in Bolivia. Indeed, this led the World Vision organisation to categorise Bolivia as one of the countries most prone to flooding  \cite{worldvision2023}. Therefore, precipitation monitoring is critical to the success of agricultural activities in rural areas. Unfortunately, current methods to track rainy seasons rely primarily on a few installed weather stations. Currently, Bolivia has fewer than 150 rain gauges, weather, and hydrological stations installed throughout the country, 60\% of which are in the main cities: La Paz, Cochabamba, and Santa Cruz. This leaves large regions of the country without precise monitoring \cite{inadhi2024}. 

Similar to other countries in the region, the shortage of advanced technologies to monitor precipitation and other weather variables in rural areas has caused the risk to go unnoticed. Heavy rainfall can cause significant damage and disruption to communities, infrastructure, and the environment. Protection methods against these events can be classified as passive or active. Passive methods are generally preventive, such as elevating structures above flood levels with dams or floodwalls, and incorporating proper drainage systems. Meanwhile, active strategies against potential heavy rainfall involve implementing weather forecasting, radar technology, or real-time monitoring systems to predict and alert communities of impending heavy rainfall events. Such methods are absent in most Latin American countries \cite{worldvision2023}. The automation of these instruments can not only increase reliability, but also improve the timely availability of data, and therefore, greatly contribute to improving agriculture and livestock, as well as to the early prevention of droughts and floods.

The growing availability of low-cost rain gauges makes data acquisition at a finer scale particularly attractive for regions with limited resources or where traditional weather radar coverage is sparse. Therefore, this project consists of developing a rainfall monitoring system composed of low-cost end devices for in situ rain measurement, incorporating sensors for temperature, soil humidity, ambient humidity, and solar radiation. Moreover, we propose an Internet of Things (IoT) system oriented toward capturing data from multiple end devices distributed throughout remote regions where the internet is unavailable. Specifically, we propose a Wide Area Network (WAN) for the system, in which each device can send precipitation and sensor data via SMS (Short Message Service) through GSM/GPRS cellular networks. We not only make this information available through a web application but also leverage it for heavy rainfall prediction. With this in mind, we propose and deploy on the website a predictor based on a Graph Neural Network (GNN), which is suitable for modelling the intrinsic features present in meteorological data from multiple locations.

The rest of this paper is organised as follows. In Section \ref{sec:related_works}, we review recent approaches for rainfall monitoring and prediction. In Section \ref{sec:proposed_system}, we present the development stages of the proposed system. In Section \ref{sec:forecast_model}, we describe the implementation of the forecasting model based on graph learning. Afterwards, Section \ref{sec:experimental_results} informs about the obtained results when predicting rainfall levels. Section \ref{sec:conclusions} concludes the paper.   

\section{Related Works}
\label{sec:related_works}

Data acquisition of precipitation measurements using IoT is a rapidly evolving field thanks to the growing availability of low-cost sensors and embedded devices. Several studies have explored the development and calibration of capacitive \cite{sa2013}, optical \cite{steele2006}, and acoustic \cite{avanzato2020} sensors. Yet, rain gauges still stand out due to their high precision and simplicity in measuring the amount of precipitation during a given period and over a specific area. Low-cost automated rain gauges can be clustered into four groups: siphon-based, wave-based, weighting-based, and tipping bucket-based \cite{tapiador2012}. The first group collects rainfall in a chamber, and empty it automatically once a certain amount of rainfall has been accumulated and registered. Wave-based gauges work by emitting ultrasonic or radar waves towards the surface of a collection chamber to calculate the distance to the surface and estimate the quantity of water. Meanwhile, weight-based rain gauges measure the weight of the collected water to determine the amount of rainfall. While these three approaches are low-cost, it is worth noting that they have capacity range limitations, which can be troublesome during heavy rain seasons \cite{urban2021}. 

\begin{figure}[htbp]
\centerline{\includegraphics[width=.5\textwidth]{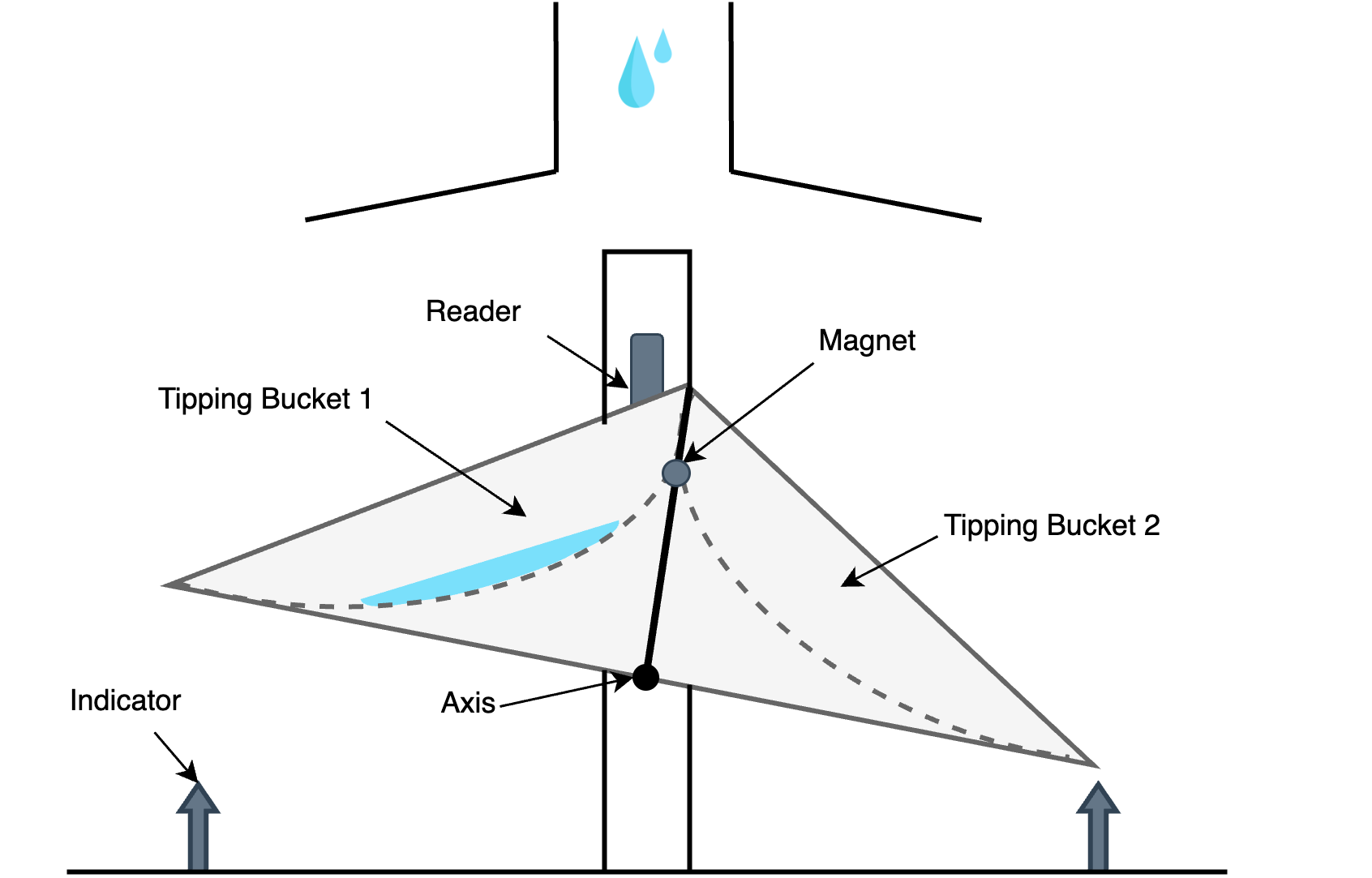}}
\caption{Tipping bucket-based mechanism to measure waterfall.}
\label{fig:tipping_bucket}
\end{figure}

Moving on, tipping-bucket-based devices record rainfall data precisely by tipping a small bucket when it fills with a specific amount of water \cite{kruger2023}, as shown in Figure \ref{fig:tipping_bucket}. Currently, the cost of tipping bucket rain gauges varies from \$100 \cite{davisinstruments2024} to \$1,600 \cite{vaisala2024}, depending on the brand, features, and accuracy. Nevertheless, their manufacturing and maintenance are accessible in terms of costs compared to weather stations. Rain gauge measurements can be combined with other rain sensors to enhance the accuracy and coverage of rainfall monitoring. This specific approach is carried out in \cite{wu2020}, where the authors fused precipitation data from the TRMM 3B42 V7 satellite and 796 rain gauges distributed in mainland China. Moreover, they merged climate data from GridSat-B1, which is a repository of global infrared measurements from multiple geostationary satellites, with the purpose of finding the spatial and temporal patterns for more accurate rainfall estimation. Similarly, researchers in \cite{yilmaz2005} combine data from radar, satellite and rain gauge sensors to tackle the coverage limitations of each data source and provide a more comprehensive hydrological forecast. 

When it comes to monitoring and prediction, Deep neural networks (DNNs), such as convolutional neural networks (CNNs) and recurrent neural networks (RNNs), have shown impressive capabilities for rainfall prediction \cite{dotse2024}. RNNs models, in particular long short-term memory (LSTM) \cite{khan2020} and gated recurrent unit (GRU) models \cite{chhetri2020}, have yielded compelling prediction outcomes when processing time series. Meanwhile, other works have centred their efforts on combining predictions from multiple machine learning (ML) algorithms \cite{hussein2022}, a technique known as ensemble learning. For example, Ahmadi et al. \cite{ahmadi2023} combined the outputs of multiple variants of a new sequential minimal optimisation (SMO) regression model using dagging (DA), random committee (RC) and additive regression (AR) to address the limitations of the model. They processed monthly rainfall data spanning 30 years from 1988 to 2018, including evaporation, maximum and minimum temperatures, maximum and minimum relative humidity, and sunshine hours in Kermanshah, Iran. Consequently, their results showed that DA-SMO outperformed other algorithms, and minimum relative humidity had the greatest effect on rainfall prediction.

Despite the progress described above, scaling up rain gauge networks to cover large geographical areas and handle real-time data processing remains a challenge. Edge computing-based solutions are being explored to minimise sensor data transmission to a central server and maximise local processing \cite{wardana2021, rojas2022}. On the other hand, Graph Neural Networks (GNNs), a special type of DNNs for processing data that can be represented as graphs, have emerged as a promising approach for rainfall prediction, leveraging the inherent graph structure of meteorological data. This not only helps reduce the need of high-quality, labeled rainfall data, but also supports developing more precise estimation models for those regions where connectivity, electricity, or computing is inaccessible. Recent research in \cite{li2023} has demonstrated the effectiveness of GNNs in capturing spatial dependencies and temporal correlations in heatwave patterns, leading to improved forecasting accuracy. 

Moving on, researchers in \cite{peng2023} introduce a novel GNN architecture that effectively integrates multi-scale features from Numerical Weather Prediction (NWP) models to enhance rainfall prediction. Similarly, \cite{chen2024} showcases the potential of GNNs in improving precipitation forecasting skills by coupling physical variables through graph structures. Furthermore, the work presented in \cite{jeon2024} demonstrates the use of GNNs to assess the impact of meteorological observations on weather forecasts. Consequently, these recent works highlight the growing interest and potential of GNNs in revolutionising rainfall prediction by effectively modelling complex spatio-temporal relationships in meteorological data. 

While rain gauges are a well-established tool for precipitation measurement, their deployment in rural regions often faces challenges due to limited infrastructure and connectivity. This scarcity of data hinders the development of accurate precipitation prediction models, particularly GNNs that leverage spatial relationships between data points. Consequently, the present proposal aims to bridge this gap by introducing a cost-effective and sustainable rain gauge network for rural areas, specifically designed to generate high-quality data suitable for training graph neural network models. This will contribute to improved precipitation prediction in data-sparse regions, with potential benefits for agriculture, water resource management, and disaster preparedness.

\section{Proposed System}
\label{sec:proposed_system}

\begin{figure}[htbp]
\centerline{\includegraphics[width=.5\textwidth]{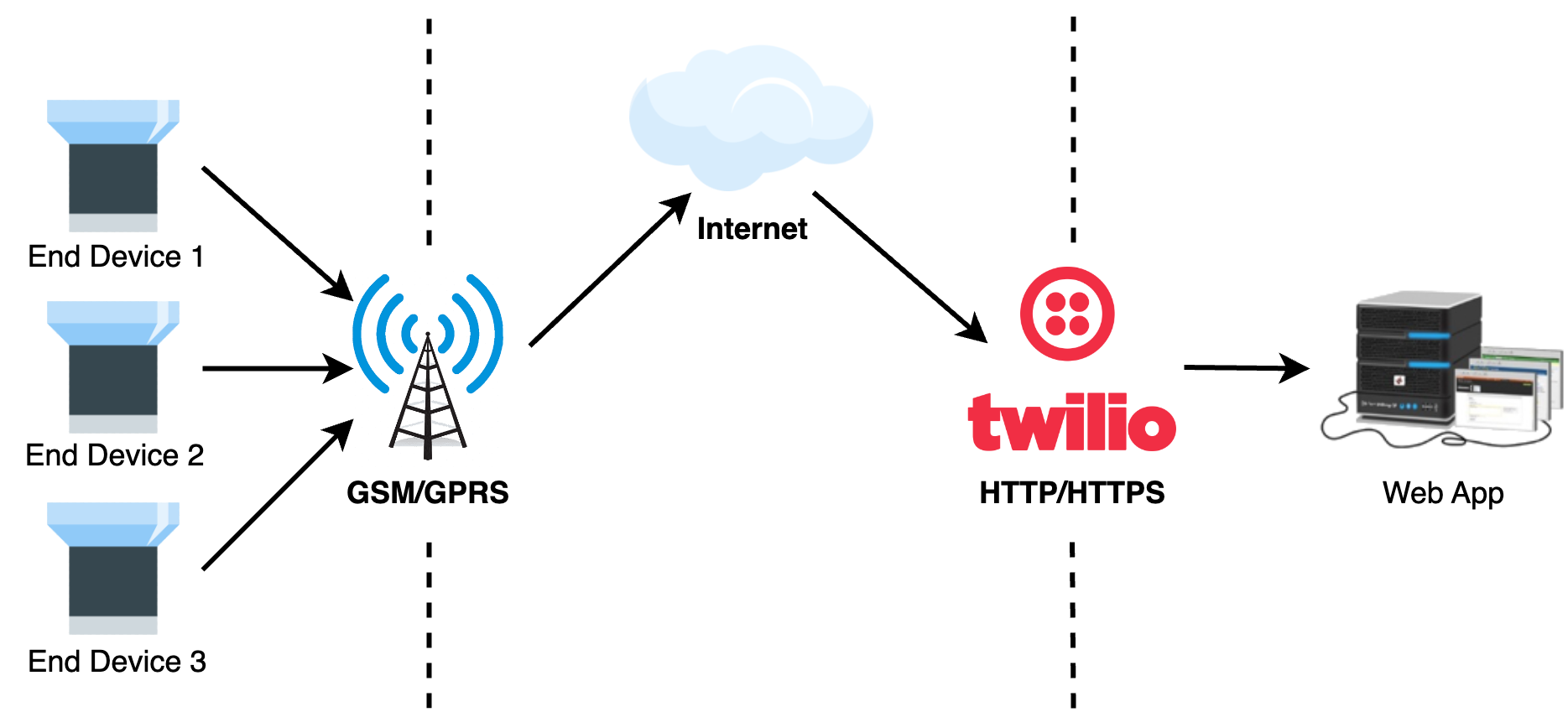}}
\caption{Proposed architecture of a rainfall data acquisition and analysis system.}
\label{fig:architecture}
\end{figure}

The architecture diagram of the proposed system is illustrated in Figure \ref{fig:architecture}, where two components can be distinguished: a set of end devices and the IoT system. 

\subsection{End Devices}

Three end devices were implemented using 3D design software SolidWorks, 3D printing, and prototyping components, with the aim of deploying the devices in remote areas. A tipping bucket rain gauge mechanism was embedded inside each of the end devices, which also contain sensors for temperature, soil humidity, ambient humidity, and solar radiation measurement. We integrated a GSM/GPRS module into a power supply mechanism composed of a 1.5 Ah battery of 7 volts and a solar panel. Each rain gauge was built based on a precision 0.2mm rocker mechanism, illustrated in Figure \ref{fig:tipping_bucket}. Each device was designed to let rainwater fall through an upper funnel onto the tipping buckets. The waterfall fills one tipping bucket at a time, and each tipping bucket can hold up to 0.2mm of rainwater. The mechanism automatically drops its contents once a tipping bucket is full, and the other bucket is positioned in front of the funnel opening to repeat the water collection process. Each time the bucket is tilted, a magnetic pulse is generated at the reader level. These pulses are counted in the Arduino Mega microcontroller to measure the precipitation. This mechanism was chosen after several unsuccessful trials with ultrasonic-based rain gauges.

\begin{figure}[t]
\centering
    \begin{minipage}{0.45\linewidth}
    \centering
    \includegraphics[width=\linewidth]{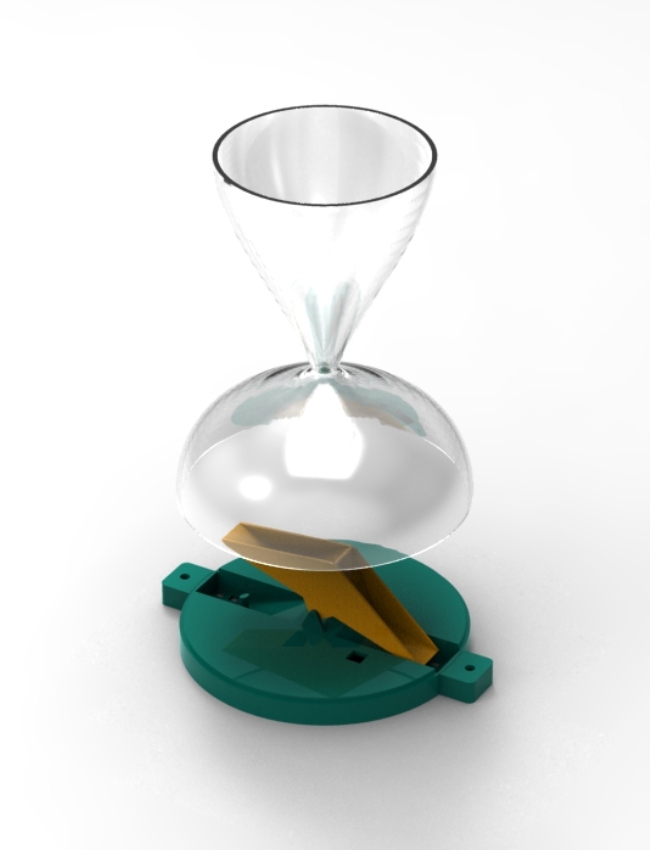}
    \label{fig:solidworks}
    \end{minipage}
    \begin{minipage}{0.45\linewidth}
        \centering
        \includegraphics[width=\linewidth]{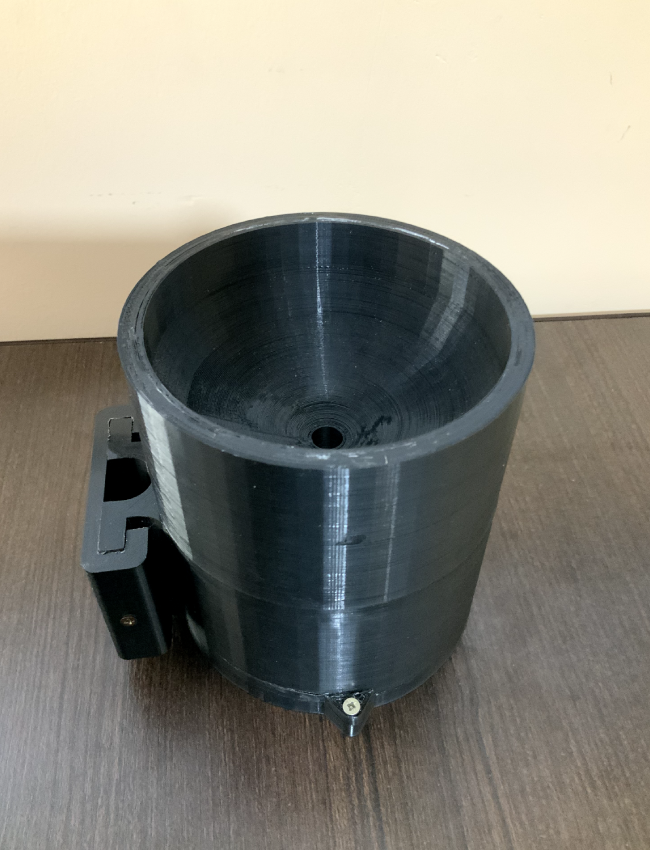}
        \label{fig:prototype}
    \end{minipage}    
    \caption{On the left, an illustration of the Solidworks (CAD) model of the tipping rain gauge integrated to the end devices. On the right, a finished rain gauge after its manufacture.}
\end{figure}

While one part of the Arduino Mega microcontroller circuit is dedicated to counting the rain gauge pulses, another part is dedicated to recording measurements from the other sensors. Specifically, we used DHT11, HL-69, and ML8511 sensors for temperature and humidity, soil moisture, and solar radiation measurement, respectively. The microcontroller then communicates with the GSM module using a USART (Universal Synchronous/Asynchronous Receiver/Transmitter) port. The USART receiver pin is enabled to generate interrupts. The transmitted and received data are stored in a buffer and then sent to the GSM module, which in turn sends SMS messages with the information formatted as a reading every 15 minutes. The remote weather station is powered by a 7V battery, and a solar panel is connected to it for charging. A lead-acid battery is used due to its rechargeable nature. The energy from the solar panel is sent to a rectifier bridge circuit and then regulated to a voltage of 5V, which can power both the microcontroller and the GSM module. In this way, the circuit has an uninterrupted power supply, either from the battery or the solar panel.

\subsection{IoT System}

The WAN used in the present research leverages GSM/GPRS cellular connectivity to deliver long-range wireless communications on the order of hundreds of kilometres. This was important not only because 3G, 4G, 5G, and LTE are not fully available yet in rural areas in Bolivia, but also because SMS sent through GSM/GPRS are currently inexpensive. Therefore, we integrated a GSM/GPRS module inside the end devices to send sensor data within SMS messages to an SMS Twilio receiver. Specifically, each microcontroller is programmed to encode sensor and precipitation measurements into strings of a maximum of 160 characters. Then, this data is separated into SMS messages to be sent to the web application every few seconds. Once an SMS is received, Twilio is programmed to send Hypertext Transfer Protocol (HTTP) requests to the web application through its application programming interface (API).

\begin{figure}[htbp]
\centerline{\includegraphics[width=.5\textwidth]{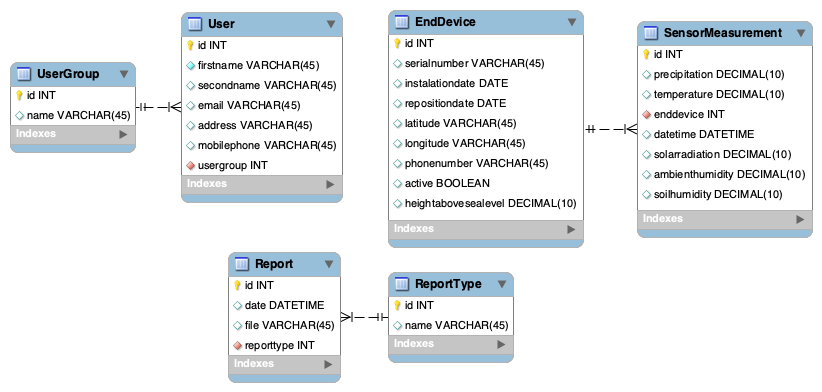}}
\caption{Entity-Relation diagram of the principal database.}
\label{fig:database}
\end{figure}

\begin{figure}[t]
\centering
    \begin{minipage}{0.90\linewidth}
        \centering
        \includegraphics[width=\linewidth]{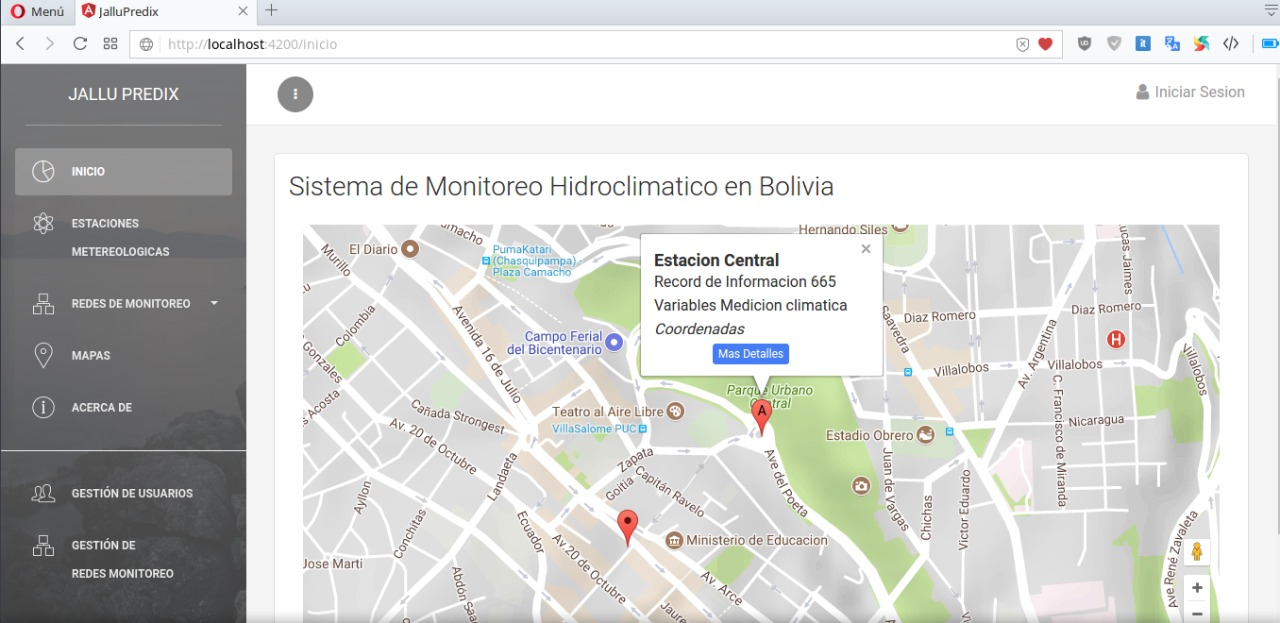}
        \label{fig:screenshot1}
        \vspace{1mm}
        
        \includegraphics[width=\linewidth]{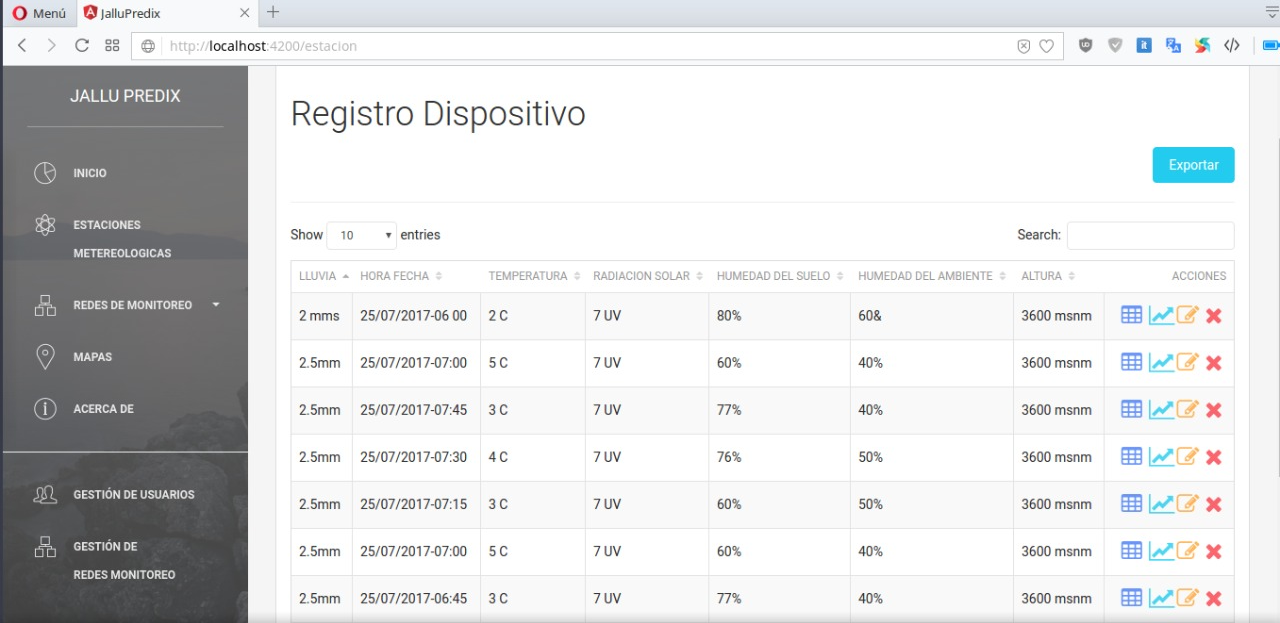}
        \label{fig:screenshot2}
        \vspace{1mm}
        
        \includegraphics[width=\linewidth]{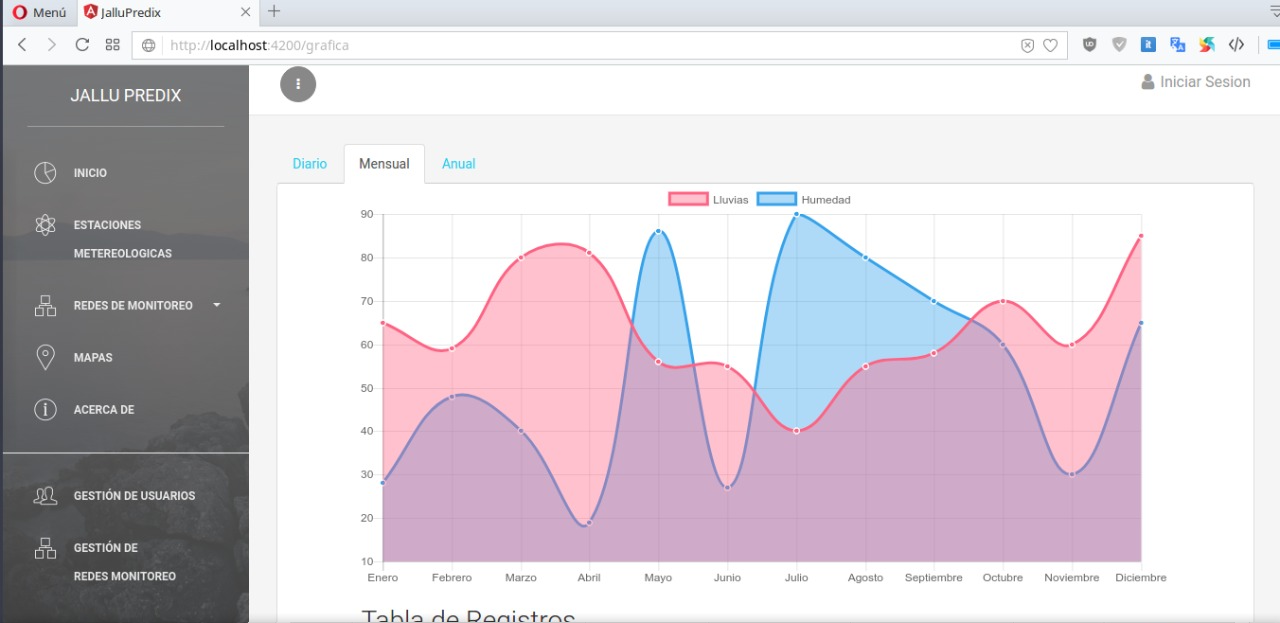}
        \label{fig:screenshot3}
    \end{minipage}
    \caption{Screenshots of the web application.}
    \label{fig:screenshots}
\end{figure}

In the central server, a web application named JalluPredix was developed using the MEAN.JS stack (MongoDB, Express, Angular, Node.JS). It contains features for the management of users, end devices, and monitoring networks. Within the web application, we implemented the database structure shown in Figure \ref{fig:database} using MongoDB. Figure \ref{fig:screenshots} shows some interfaces designed for end-users to interact with the monitoring system.

\section{Forecast Model: Graph Neural Network}
\label{sec:forecast_model}

\subsection{Data Collection and Processing}
\label{sec:data}

Once the Twilio SMS receiver forwards an SMS to the server as an HTTP request, an NGINX server receives the request and communicates it to a REST API, which, in turn, saves the sensor data within a MongoDB database. This approach has been tested for a couple of years, yielding precipitation data about a specific region in La Paz, Bolivia. However, the present proposal led us to search for secondary data sources to better model a graph-based architecture based on historical data. Thus, we requested daily precipitation data at a local scale (municipalities and departments) from the National Service of Meteorology and Hydrology of Bolivia (SENAMHI). Nevertheless, this department required a formal letter to request the information. Since writing a letter every time we need to update the proposed monitoring system is impractical, we scraped the precipitation data from SENAMHI's website using Python, Selenium, and BeautifulSoup. It is worth mentioning that this data is publicly available for users through the INADHI section of the website \cite{inadhi2024}, but the format is not user-friendly and requires filling in several fields before downloading an Excel file with the requested measurements. 

\begin{figure}[htbp]
\centerline{\includegraphics[width=.5\textwidth]{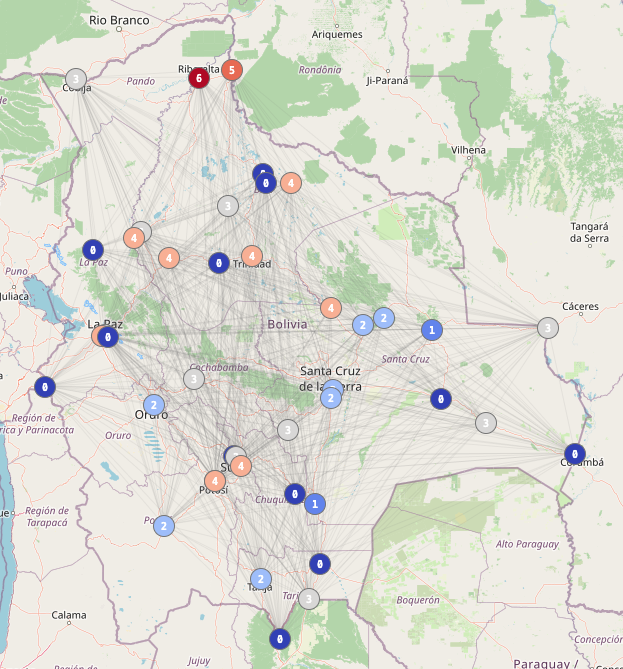}}
\caption{ Geographic locations of the 41 selected weather stations; Filled color shows the annual average number of heavy rainfalls events at each node.}
\label{fig:map}
\end{figure}

The scraped data consisted of daily precipitation measurements from 1 January 2017 to 30 April 2024. However, it required removing weather stations with too many missing values. We also removed unwanted characters present in the records (e.g., measurement units, commas, spaces). Moreover, the metadata for each end device (i.e., altitude, geolocation, type, sub-type, among other details) was obtained manually and helped to structure the graphs. A plot showing the locations of the resulting 41 weather stations and their average number of heavy rainfalls per year (precipitation per day $\geq$ 8mm) is shown in Figure \ref{fig:map}.

\subsection{Model Architecture}

In this work, we approached the heavy rainfall prediction problem from a graph perspective, noting the space and altitude relations between multiple rainfall measurements. Therefore, we propose the application of a GNN to understand historical measurements and their relations for rainfall precipitation prediction based on data from multiple weather stations or rain gauges. The proposed GNN models the weather stations as nodes in a graph, and the edges between nodes represent distance relationships. Therefore, the GNN learns how information flows through the graph to predict rainfall in different sections across the country. We can note the graph elements as: 

\begin{itemize}
    \item \textbf{Nodes $(V)$:} Each node $v_i \in V$ represents a weather station.
    \item \textbf{Edges $(E)$:} Each edge $(v_i, v_j) \in E$ indicates the distance relationship between stations $i$ and $j$.
    \item \textbf{Node Features $(X)$:} Each node $v_i$ has a feature vector $x_i \in X$, which contains meteorological data from that station (mainly past rainfall).
    \item \textbf{Adjacency Matrix (A):} This matrix $A \in \mathbb{R}^{|V|x|V|}$ defines the graph structure, where $A_ij = d$ and the $d$ accounts for the distance between the nodes $i$ and $j$. 
\end{itemize}

The core operation of the GNN is the graph convolution layer (GCL), which updates node features based on information from neighboring nodes, weighted by the adjacency matrix, and then applies a linear transformation and activation function (ReLu). Then, GCL can be defined as in Equation \ref{eq:core_operation}:

\begin{equation}
H^{l+1} = \sigma(D^{-1/2} A D^{-1/2} H^{l} W^{l})\label{eq:core_operation}
\end{equation}

Where:
\begin{itemize}
    \item $H^{l}$ is the matrix of node features at layer $l$ (${H^{0} = X}$).
    \item $W^{l}$ is the weight matrix for layer $l$.
    \item $D$ is the diagonal degree matrix ($D_{ii}$ is the sum of weights for edges connected to node i).
    \item $\sigma$ is the activation function ReLU.
\end{itemize}

After four GCLs, the final node features $H^{L}$ are used to make rainfall predictions. Then, we include a fully connected layer followed by a regression layer to predict the amount of rainfall for each weather station. If rainfall is predicted to surpass 8mm, we flag a potential heavy rainfall event, which allows us to notify the authorities or community before the adversity. Finally, each resulting model is evaluated using the mean squared error (MSE), mean absolute error (MAE) and Pearson correlation coefficient (r), as defined in Equations \ref{eq:mse}, \ref{eq:mae}, and \ref{eq:r}, respectively. 

\begin{equation}
MSE = (\frac{1}{n})\sum_{i=1}^{n}\left ( y_{i} - \hat y_{i} \right )^2
\label{eq:mse}
\end{equation}

\begin{equation}
MAE = (\frac{1}{n})\sum_{i=1}^{n}\left | y_{i} - \hat y_{i} \right |
\label{eq:mae}
\end{equation}

\begin{equation}
r = \frac{cov(X,\ y)}{\sigma(X_i).\sigma(y_i)}
\label{eq:r}
\end{equation}

\section{Experimental Results}
\label{sec:experimental_results}

To evaluate the performance of our proposed GNN model for heavy rainfall prediction, we conducted experiments on the dataset described in Section \ref{sec:data}, which comprises meteorological data from 41 weather stations across Bolivia. The dataset spanned 72 months, with daily measurements of historical rainfall data. We pre-processed the data by standardising features and addressing missing values through interpolation. We constructed the GNN and the graph structure using TensorFlow, considering the spatial proximity between stations. The adjacency matrix was weighted based on the inverse distance between stations, emphasising stronger connections for closer stations. We split the dataset into training (70\%), validation (20\%), and testing (10\%) sets. The GNN model was trained using the Adam optimiser with a learning rate of 0.01 and a batch size of 64. Early stopping was employed to prevent overfitting. The model's performance is shown in Table \ref{tab:results}.

\color{black}

\begin{table}[htbp]
\caption{Performance metrics of the tested GNN models.}
\begin{center}
\begin{tabular}{|c|c|c|c|c|}
\hline

\textbf{\makecell{Model\\ Name}} & Layers & \textbf{MSE} & \textbf{MAE} & \textbf{r} \\
\hline
A & GCL10-GCL10-GCL10-GCL10 & 16.344 & 9.512& 5.891 \\
B & GCL16-GCL32-GCL64-GCL128 & 17.462 & 10.261 & 6.002 \\
C & GCL128-GCL64-GCL32 & 18.756 & 10.889 & 7.204 \\
D & GCL32-GCL64-GCL128 & 19.112 & 11.993 & 8.771 \\
\hline

\end{tabular}
\label{tab:results}
\end{center}
\end{table}

\begin{figure}[htbp]
\centerline{\includegraphics[width=.45\textwidth]{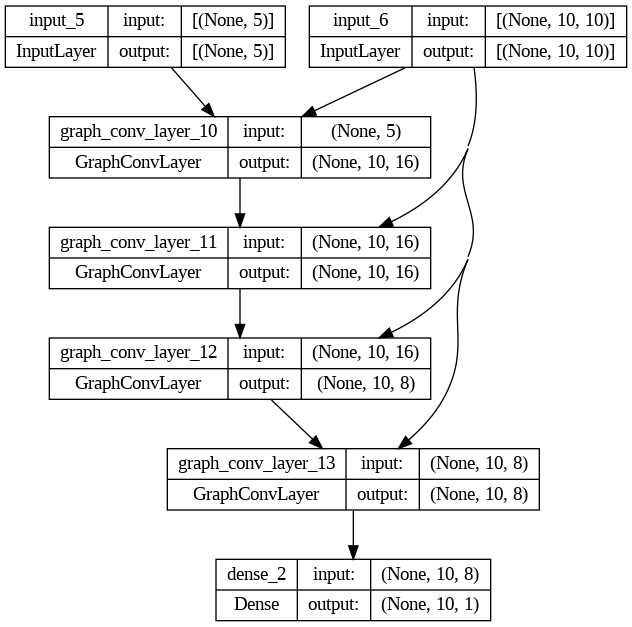}}
\caption{Best performing GNN model.}
\label{fig:gnn}
\end{figure}

The best-performing GNN model, model A (shown in Figure \ref{fig:gnn}), demonstrated promising results, achieving an MSE of 16.344, an MAE of 9.512, and an r of 5.891 on the testing set. While all models shown in Table \ref{tab:results} effectively captured spatial dependencies and leveraged meteorological features to accurately predict rainfall, model A outperformed the others in terms of accuracy metrics.

\color{black}

\section{Conclusions}
\label{sec:conclusions}

This paper introduces an automated system for monitoring and predicting heavy rainfall. This system offers a more reliable and accurate alternative based on low-cost rain gauges, particularly beneficial in remote areas and towns with limited road access or internet connectivity. The developed meteorological station is easily adaptable, capable of measuring additional variables such as wind, evaporation, and atmospheric pressure with the integration of suitable sensors. This information could be seamlessly incorporated into SMS messages transmitted via the GSM/GPRS network and processed by a central web system.

Further development is required in the data collection and processing stages to consider all relevant variables (temperature, soil and air humidity, radiation, precipitation) for better predictions. The present project focuses on proposing low-cost rain measurement devices that could enable the correlation of data from various end devices and allow us to forecast precipitation in regions without installed rain gauges. Ultimately, the collected information has wider applications beyond agriculture, benefiting society as a whole, for instance, in predicting and mitigating natural disasters.

While the cost of our prototype ranges from \$280 to \$320, its design is primarily for research purposes and will be refined in our future research. The next phase will involve utilising industrial electronic components to create more durable and higher-quality devices. Moreover, future research will focus on embedding the GNN model within the end devices to provide localised precipitation predictions based on historical data from neighbouring devices.

\section*{Acknowledgement}
The preliminary work of this research won the Hackathon ``Mi Madre Tierra'' \cite{jallupredix2016}, which was organised by the Ministry of Environment and Water in Bolivia in November 2016. Subsequent research was funded in part by the Ministry (grant reference MMAYA/CIP/005/2017) and Google AI.

\bibliographystyle{ieeetr}
\bibliography{bibliography}
\end{document}